\documentclass[11pt,a4paper]{article}
\usepackage[hyperref]{acl2019}
\usepackage{times}
\usepackage{latexsym}
\usepackage{multirow}

\usepackage{graphicx}
\usepackage{siunitx}
\usepackage{booktabs}
\usepackage{amsmath}
\usepackage{amsfonts}
\usepackage{enumitem}

\usepackage{url}

\usepackage{cleveref}
\crefformat{section}{\S#2#1#3}
\usepackage{color}
\usepackage[textsize=scriptsize]{todonotes}
\definecolor{blue}{RGB}{0, 93, 170}			
\definecolor{darkgreen}{RGB}{0, 102, 0}

\newcommand{\bo}{\mathbf{o}}
\newcommand{\ba}{\mathbf{a}}
\newcommand{\bh}{\mathbf{h}}
\newcommand{\bz}{\mathbf{z}}
\newcommand{\rW}{\mathrm{W}}

\aclfinalcopy 

\title{Extracting Multiple-Relations in One-Pass \\ with Pre-Trained Transformers}
\author{
 Haoyu Wang\thanks{\, Equal contributions from the corresponding authors: \texttt{\{wanghaoy,mingtan,yum\}@us.ibm.com}. Part of work was done when Kun was at IBM.} $^\dagger$ \quad \quad \   Ming Tan$^{*\dagger}$\quad \quad Mo Yu$^{*\ddagger}$\quad \quad Shiyu Chang$^\ddagger$ \\ 
 \textbf{Dakuo Wang}$^\ddagger$\quad \quad \textbf{Kun Xu}$^\mathsection$ \quad \quad \textbf{Xiaoxiao Guo}$^\ddagger$\quad \quad \textbf{Saloni Potdar}$^\dagger$\\
 $^\dagger$IBM Watson \quad \quad $^\ddagger$IBM Research \quad \quad $^\mathsection$Tencent AI Lab\\
}

\date{}

\begin{document}
\maketitle
\begin{abstract}
The state-of-the-art solutions for extracting multiple entity-relations from an input paragraph always require a multiple-pass encoding on the input.
This paper proposes a new solution that can complete the multiple entity-relations extraction task with only \textbf{one-pass encoding} on the input corpus, and achieve a \textbf{new state-of-the-art accuracy performance}, as demonstrated in the ACE 2005 benchmark. Our solution is built on top of the pre-trained self-attentive models (Transformer). Since our method uses a single-pass to compute all relations at once, it scales to larger datasets easily; which makes it more usable in real-world applications. \footnote{\scriptsize{\url{https://github.com/helloeve/mre-in-one-pass}.}}

\end{abstract}

\section{Introduction}
Relation extraction (RE) aims to find the semantic relation between a pair of entity mentions from an input paragraph. A solution to this task is essential for many downstream NLP applications such as automatic knowledge-base completion \cite{surdeanu2012multi,riedel2013relation,verga2016multilingual},  knowledge base question answering \cite{yih2015semantic,xu2016question,yu2017improved}, and symbolic approaches for visual question answering~\cite{mao2019neuro,hu2019language}, \emph{etc}. 

One particular type of the RE task is \emph{multiple-relations extraction (MRE)} that aims to recognize relations of multiple pairs of entity mentions from an input paragraph. Because in real-world applications, whose input paragraphs dominantly contain multiple pairs of entities, an efficient and effective solution for MRE has more important and more practical implications. However, nearly all existing approaches for MRE tasks \cite{qu2014senti,gormley2015improved,nguyen2015combining} adopt some variations of the single-relation extraction (SRE) approach, which treats each pair of entity mentions as an independent instance, and requires multiple passes of encoding for the multiple pairs of entities.  The drawback of this approach is obvious -- it is computationally expensive and this issue becomes more severe when the input paragraph is large, making this solution impossible to implement when the encoding step involves deep models.

\begin{figure}[t!]
\vspace{-2mm}
  \includegraphics[width=0.45\textwidth]{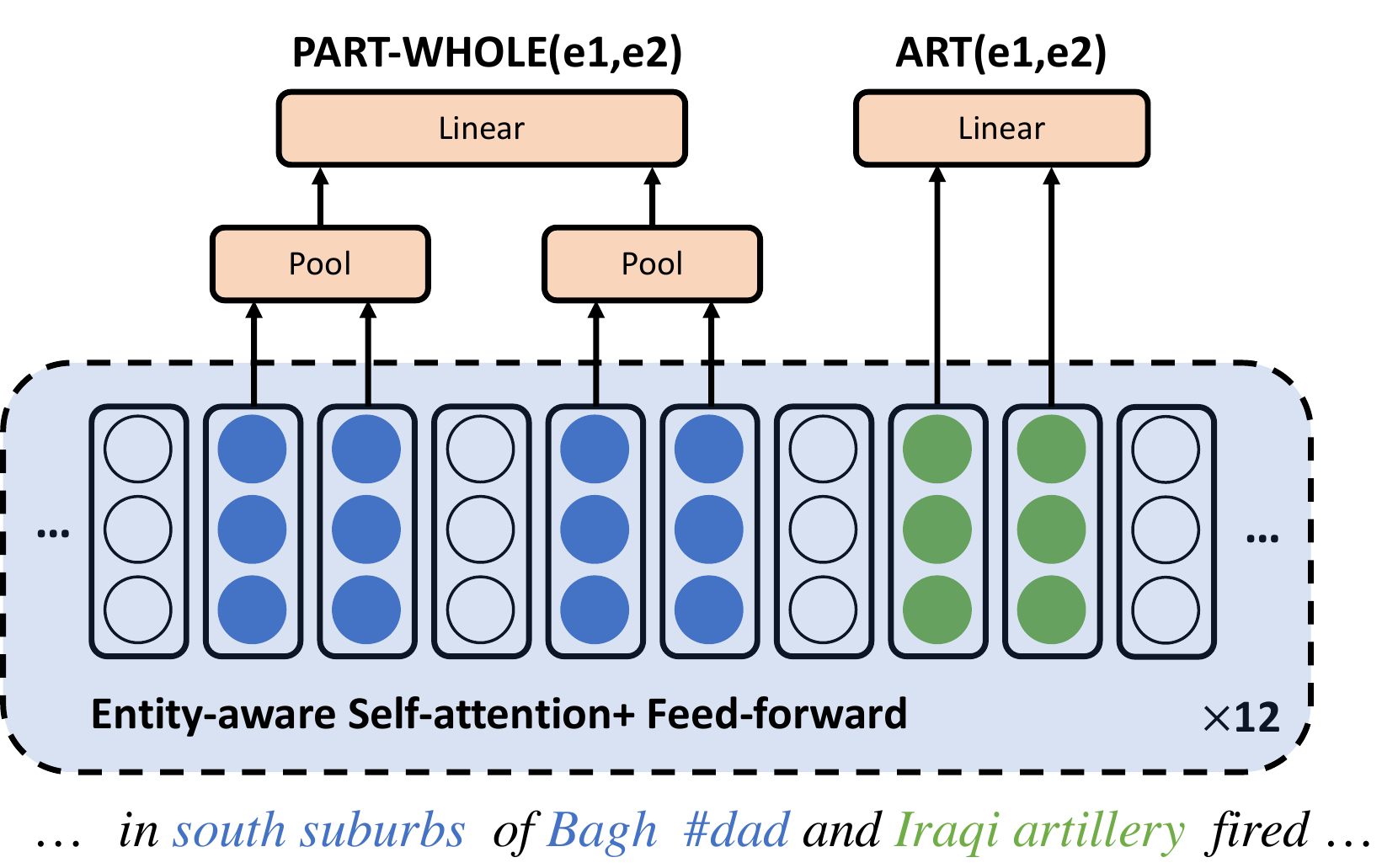}
  \caption{\small Model Architecture.  Different pairs of entities, e.g., (\emph{Iraqi} and \emph{artillery}), (\emph{southern suburbs}, \emph{Baghdad}) are predicted simultaneously.
}
  \label{fig:model-overview}
  \vspace{-3mm}
\end{figure}

This work presents a solution that can resolve the inefficient multiple-passes issue of existing solutions for MRE by encoding the input only once, which significantly increases the efficiency and scalability. Specifically, the proposed solution is built on top of the existing transformer-based, pre-trained general-purposed language encoders. In this paper we use \emph{Bidirectional Encoder Representations from Transformers (BERT)} \cite{devlin2018bert} as the transformer-based encoder, but this solution is not limited to using BERT alone. The two novel modifications to the original BERT architecture are: (1) we introduce a structured prediction layer for predicting multiple relations for different entity pairs; and (2) we make the self-attention layers aware of the positions of all entities in the input paragraph. To the best of our knowledge, this work is the first promising solution that can solve MRE tasks with such high efficiency (encoding the input in one-pass) and effectiveness (achieve a new state-of-the-art performance), as proved on the ACE 2005 benchmark.  

\section{Background}
MRE is an important task as it is an essential prior step for many downstream tasks such as automatic knowledge-base completion and question-answering. 
Popular MRE benchmarks  
include ACE~\cite{walker2006ace} and ERE~\cite{ldc2013ere}.   In MRE, given as a text paragraph  $\mathbf{x}=\{x_1, \dots, x_N\}$ and $M$ mentions $\mathbf{e}=\{e_1, \dots, e_M\}$ as input, the goal is to predict the relation $r_{ij}$ for each mention pair $(e_i, e_j)$ either belongs to one class of a list of pre-defined relations $\mathcal{R}$ or falls into a special class \emph{NA} indicating no relation.  
This paper uses ``entity mention'', ``mention'' and ``entity'' interchangeably.  

Existing MRE approaches are based on either feature and model architecture selection techniques \cite{xu2015semantic,gormley2015improved, nguyen2015combining,petroni2015core,TUD-CS-2017-0119,song2018n}, or domain adaptations approaches \cite{fu2017domain,shi2018genre}. But these approaches require multiple passes of encoding over the paragraph, as they treat a MRE task as multiple passes of a SRE task.

\label{sec:proposed_approach}
\section{Proposed Approach}
This section describes the proposed one-pass encoding MRE solution.
The solution is built upon BERT with a structured prediction layer to enable BERT to predict multiple relations with one-pass encoding, and an entity-aware self-attention mechanism to infuse the relational information with regard to multiple entities at each layer of hidden states.  The framework is illustrated in Figure \ref{fig:model-overview}. It is worth mentioning that our solution can easily use other transformer-based encoders besides BERT, \emph{e.g.} \cite{radford2018improving}.  

\subsection{Structured Prediction with BERT for MRE}
\label{ssec:prediction}
The BERT model has been successfully applied to various NLP tasks. However, the final prediction layers used in the original model is not applicable to MRE tasks.  The MRE task essentially requires to perform edge predictions over a graph with entities as nodes.  Inspired by \cite{dozat2018simpler,ahmad2018near}, we propose that we can first encode the input paragraph using BERT. Thus, the representation for a pair of entity mentions $(e_i, e_j)$  can be denoted as $\bo_i$ and $\bo_j$ respectively. In the case of a mention $e_i$ consist of multiple hidden states (due to the byte pair encoding),  $\bo_i$ is aggregated via average-pooling over the hidden states of the corresponding tokens in the last BERT layer.   We then concatenate $\bo_i$ and $\bo_j$ denoted as $[\bo_i:\bo_j]$, and pass it to a linear classifier\footnote{We also tried to use MLP and Biaff instead of the linear layer for the classification, which do not show better performance compared to the linear classier, as shown in the experiment section. We hypothesize that this is because the embeddings learned from BERT are powerful enough for linear classifiers. Further experiments is needed to verify this.} to predict the relation
\begin{equation}
\small
P(r_{ij}| \mathbf{x}, e_i, e_j)=\textrm{softmax}(\textrm{W}^L[\bo_i:\bo_j] + \mathbf{b}),
\label{pred-layer}
\end{equation}
where $\textrm{W}^L \in \mathbb{R}^{2d_z \times l}$. $d_z$ is the dimension of BERT embedding at each token position, and $l$ is the number of relation labels.

\subsection{Entity-Aware Self-Attention based on Relative Distance}
\label{ssec:attention}
This section describes how we encode multiple-relations information into the model.  The key concept is to use the relative distances between words and entities to encode the positional information for each entity.  This information is propagated through different layers via attention computations.  Following \cite{shaw2018self}, for each pair of word tokens $(x_i, x_j)$ with the input representations from the previous layer as $\bh_i$ and $\bh_j$,  we extend the computation of self-attention $\bz_i$ as:
\begin{equation}
\small
  \bz_i = \sum_{j=1}^N \frac{\exp e_{ij}}{\sum_{k=1}^N \exp e_{ik}} (\bh_j \rW^V + \ba_{ij}^V),
\label{value-entity}
\end{equation}
\begin{equation}
\small
 \text{where}\quad e_{ij} = \bh_i \rW^Q(\bh_j \rW^K + \ba^K_{ij})/\sqrt{d_z}.
\label{attention-entity}
\end{equation}

$\rW^Q, \rW^K, \rW^V \in \mathbb{R}^{d_z \times d_z}$ are the parameters of the model, and $d_z$ is the dimension of the output from the self-attention layer.  

Compared to standard BERT's self-attention, $\ba_{ij}^V, \ba_{ij}^K \in \mathbb{R}^{d_z}$ are extra, which could be viewed as the edge representation between the input element $x_i$ and $x_j$ . Specifically, we 
devise $\ba_{ij}^V$ and $\ba_{ij}^K$ 
to encourage each token to be aware of the relative distance to different entity mentions, and vice versa.

Adapted from \cite{shaw2018self}, we argue that the relative distance information will not help if the distance is beyond a certain threshold. Hence we first define the distance function as:
\begin{equation}
\small
\begin{aligned}
d(i,j) &= min(max(-k, (i - j)), k).
\label{eq:distance}
\end{aligned}
\end{equation}
This distance definition clips all distances to a region [$-k$, $k$].
$k$ is a hyper-parameter to be tuned on the development set. We can now define $\ba_{ij}^V$ and $\ba_{ij}^K$ formally as:
\begin{equation}
\small
\begin{aligned}
\ba_{ij}^V, \ba_{ij}^K &= 
\begin{cases}
w_{d(i,j)}^V, w_{d(i,j)}^K, & \text{if } x_i \in e \\
w_{d(j,i)}^V, w_{d(j,i)}^K, & \text{if } x_j \in e \\
\bf 0, & \text{else}.
\end{cases}
\end{aligned}
\end{equation}

\begin{figure}[t!]
  \centering
  \includegraphics[width=0.36\textwidth]{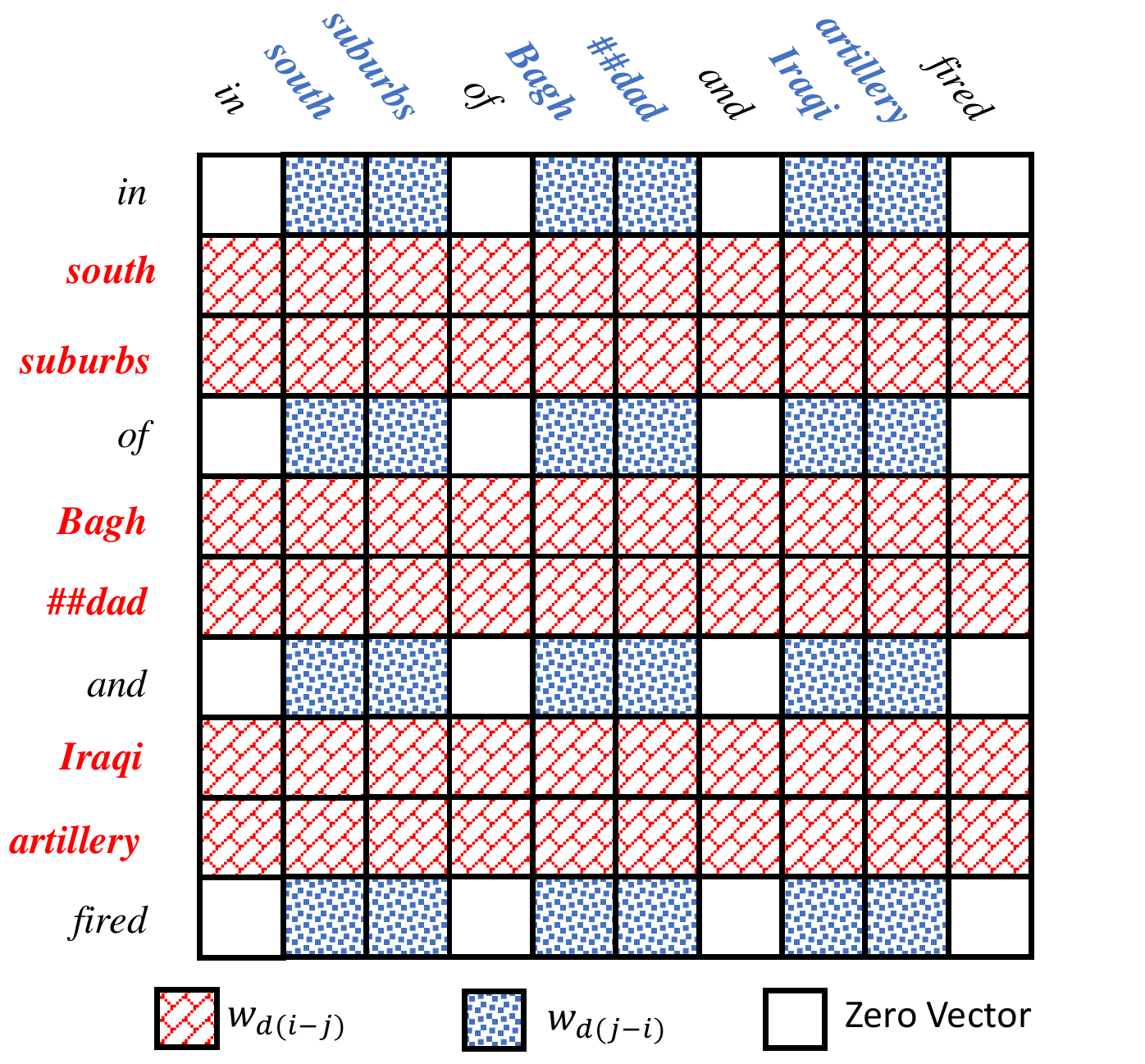}
  \caption{\small{Illustration of the tensor $\{\ba_{ij}^K\}$ introduced in self-attention computation. Each red cell embedding is defined by $w_{d(i-j)}$, as the distance from entity $x_i$ to token $x_j$. Each blue cell embedding is defined by $w_{d(j-i)}$, as the distance from the entity $x_j$ to token $x_i$ .  White cells are zero embeddings since neither $x_i$ nor $x_j$ is entity. The $\{\ba_{ij}^V\}$ follows the same pattern with independent parameters.}}
  \label{fig:entity-attention}
  \vspace{-2mm}
\end{figure}

As defined above, if either token $x_i$ or $x_j$ belongs to an entity, we will introduce a relative positional representation according to their distance. The distance is defined in an entity-centric way as we always compute the distance from the entity mention to the other token. If neither $x_i$ nor $x_j$ are entity mentions, we explicitly assign a zero vector to $\ba_{ij}^K$ and $\ba_{ij}^V$. When both $x_i$ and $x_j$ are inside entity mentions, we take the distance as $d(i,j)$ to make row-wise attention computation coherent as depicted in Figure \ref{fig:entity-attention}.

During the model fine-tuning, the newly introduced parameters $\{w_{-k}^K,...,w_k^K\}$ and $\{w_{-k}^V,...,w_k^V\}$  are trained from scratch. 

\section{Experiments}

\begin{table*}[!htbp]
\centering
\small
\begin{tabular}{lccccc}
\toprule
\textbf{Method}                                                                                           & \textbf{dev} & \textbf{bc}    & \textbf{cts}   & \textbf{wl}    & \textbf{avg}   \\ \midrule
\multicolumn{6}{c}{\underline{\emph{Baselines w/o Domain Adaptation (Single-Relation per Pass)}}} \\ 
Hybrid FCM \cite{gormley2015improved}                                                       & -            & 63.48          & 56.12          & 55.17          & 58.26          \\ 
Best published results w/o DA (from Fu et al.) & -            & 64.44          & 54.58          & 57.02          & 58.68          \\ 
BERT fine-tuning out-of-box                                                                                           & 3.66         & 5.56           & 5.53           & 1.67           & 4.25           \\ 
\midrule
\multicolumn{6}{c}{\underline{\emph{Baselines w/ Domain Adaptation (Single-Relation per Pass)}}} \\ 
Domain Adversarial Network \cite{fu2017domain}  &-& 65.16          & 55.55          & 57.19          & 59.30          \\ 
Genre Separation Network \cite{shi2018genre}                & -            & 66.38          & 57.92          & 56.84          & 60.38          \\ 
\midrule
\multicolumn{6}{c}{\underline{\emph{Multi-Relation per Pass}}} \\ 
BERT$_{\textrm{SP}}$ (our model in \cref{ssec:prediction}) & 64.42        & 67.09          & 53.20          & 52.73          & 57.67          \\ 
Entity-Aware BERT$_{\textrm{SP}}$ (our full model) & \bf 67.46        & \bf 69.25          & \bf 61.70          & \textbf{58.48} & \textbf{63.14} \\
BERT$_{\textrm{SP}}$ w/ entity-indicator on input-layer &   65.32      &      66.86     &     57.65      &     53.56      &     59.36      \\ 
BERT$_{\textrm{SP}}$ w/ pos-emb on final att-layer                           & 67.23        & 69.13          & 58.68          & 55.04          & 60.95          \\
\midrule
\multicolumn{6}{c}{\underline{\emph{Single-Relation per Pass}}} \\ 
BERT$_{\textrm{SP}}$ (our model in \cref{ssec:prediction}) & 65.13 & 66.95 & 55.43 & 54.39 & 58.92 \\
Entity-Aware BERT$_{\textrm{SP}}$ (our full model)                                                    & 68.90        & 68.52          & {63.71} & 57.20          & 63.14          \\ 
BERT$_{\textrm{SP}}$ w/ entity-indicator on input-layer & 67.12        & 69.76          & 58.05          & 56.27          & 61.36          \\
\bottomrule
\end{tabular}
\vspace{-2mm}
\caption{Main Results on ACE 2005.}
\vspace{-2mm}
\label{tab:ace}
\end{table*}


We demonstrate the advantage of our method on a popular MRE benchmark, ACE 2005~\cite{walker2006ace}, and a more recent MRE benchmark, SemEval 2018 Task 7~\cite{gabor2018semeval}.
We also evaluate on a commonly used SRE benchmark SemEval 2010 task 8~\cite{hendrickx2009semeval}, and achieve state-of-the-art performance.
\subsection{Settings}

\paragraph{Data} For \textbf{ACE 2005}, we adopt the multi-domain setting and split the data following \cite{gormley2015improved}:
we train on the union of news domain (nw and bn), tune hyperparameters on half of the broadcast conversation (bc) domain, and evaluate on the remainder of broadcast conversation (bc), the telephone speech (cts), usenet newsgroups (un), and weblogs (wl) domains.
For \textbf{SemEval 2018 Task 7}, we evaluate on its sub-task 1.1.
We use the same data split in the shared task. 
The passages in this task is usually much longer compared to ACE. Therefore we adopt the following pre-processing step -- for the entity pair in each relation, we assume the tokens related to their relation labeling are always within a range from the fifth token ahead of the pair to the fifth token after it. Therefore, the tokens in the original passage that are not covered by the range of ANY input relations, will be removed from the input. 

\paragraph{Methods}
We compare our solution with previous works that predict a single relation per pass~\cite{gormley2015improved,nguyen2015combining,fu2017domain,shi2018genre}, our model that predicts single relation per pass for MRE, and with the following naive modifications of BERT that could achieve MRE in one-pass.


\noindent $\bullet$ \textbf{BERT$_{\textrm{SP}}$}: {BERT with structured prediction only}, which includes proposed improvement in \ref{ssec:prediction}.

\noindent $\bullet$ \textbf{Entity-Aware BERT$_{\textrm{SP}}$}: our full model, which includes both improvements in \cref{ssec:prediction} and \cref{ssec:attention}.

\noindent $\bullet$ \textbf{BERT$_{\textrm{SP}}$ with position embedding on the final attention layer}. This is a more straightforward way to achieve MRE in one-pass derived from previous works using position embeddings \cite{nguyen2015combining,fu2017domain,shi2018genre}.
In this method, the BERT model encode the paragraph to the last attention-layer. Then, for each entity pair, it takes the hidden states, adds the relative position embeddings corresponding to the target entities, and finally makes the relation prediction for this pair.

\noindent $\bullet$ \textbf{BERT$_{\textrm{SP}}$ with entity indicators on input layer}: it replaces our structured attention layer, and adds indicators of entities (transformed to embeddings) directly to each token's word embedding\footnote{Note the usage of relative position embeddings does not work for one-pass MRE, since each word corresponds to a varying number of position embedding vectors. Summing up the vectors confuses this information. It works for the single-relation per pass setting, but the performance lags behind using only indicators of the two target entities.}. This method is an extension of \cite{verga2018re} to the MRE scenario. 

\paragraph{Hyperparameters}
For our experiments, most model hyperparameters are the same as in pre-training. We tune the training epochs and the new hyperparameter $k$ (in Eq. \ref{eq:distance}) on the development set of ACE 2005.
Since the SemEval task has no development set, we use the best hyperparameters selected on ACE. For the number of training epochs, we make the model pass similar number of training instances as in ACE 2005.


\subsection{Results on ACE 2005}

\paragraph{Main Results}

Table \ref{tab:ace} gives the overall results on ACE 2005. The first observation is that our model architecture achieves much better results compared to the previous state-of-the-art methods. Note that our method was not designed for domain adaptation, it still outperforms those methods with domain adaptation. This result further demonstrates its effectiveness.

Among all the BERT-based approaches, fine-tuning the off-the-shelf BERT does not give a satisfying result, because the sentence embeddings cannot distinguish different entity pairs.
The simpler version of our approach, BERT$_{\textrm{SP}}$, can successfully adapt the pre-trained BERT to the MRE task, and achieves comparable performance at the prior state-of-the-art level of the methods without domain adaptation.

Our full model, with the structured fine-tuning of attention layers, brings further improvement of about 5.5\%, in the MRE one-pass setting, and achieves a new state-of-the-art performance when compared to the methods with domain adaptation. It also beats the other two methods on BERT in Multi-Relation per Pass.

\paragraph{Performance Gap between MRE in One-Pass and Multi-Pass}
The MRE-in-one-pass models can also be used to train and test with one entity pair per pass (\emph{Single-Relation per Pass} results in Table \ref{tab:ace}).
Therefore, we compare the same methods when applied to the multi-relation and single-relation settings.
For BERT$_\textrm{SP}$ with entity indicators on inputs, it is expected to perform slightly better in the single-relation setting, because of the mixture of information from multiple pairs. A 2\% gap is observed as expected. 
By comparison, our full model has a much smaller performance gap between two different settings (and no consistent performance drop over different domains).

The BERT$_\textrm{SP}$ is not expected to have a gap as shown in the table.
For BERT$_\textrm{SP}$ with position embeddings on the final attention layer, we train the model in the single-relation setting and test with two different settings, so the results are the same.

\paragraph{Training and Inference Time}
Through our experiment,\footnote{All evaluations were done on a single Tesla K80 GPU.} we verify that the full model with MRE is significantly faster compared to all other methods for both training and inference. The training time for full model with MRE is 3.5x faster than it with SRE. As for inference speed, the former could reach 126 relation per second compared the later at 23 relation per second. It is also much faster when compared to the second best performing approach, \emph{BERT$_{\textrm{SP}}$ w/ pos-emb on final att-layer}, which is at 76 relation per second, as it runs the last layer for every entity pair.

\paragraph{Prediction Module Selection}
Table \ref{tab:prediction_ablation} evaluates the usage of different prediction layers, including replacing our linear layer in Eq.(\ref{pred-layer}) with MLP or Biaff. Results show that the usage of the linear predictor gives better results. This is consistent with the motivation of the pre-trained encoders: by unsupervised pre-training the encoders are expected to be sufficiently powerful thus adding more complex layers on top does not improve the capacity but leads to more free parameters and higher risk of over-fitting.

\begin{table}[!t]
\small
\centering
\begin{tabular}{lccccc}
\toprule
\textbf{Method}                                      & \textbf{dev} & \textbf{bc} & \textbf{cts} & \textbf{wl} & \textbf{avg} \\ \midrule
Linear            & \bf 67.46        & \bf 69.25          & \bf 61.70          & \textbf{58.48} & \textbf{63.14} \\
MLP &    67.16 & 68.52 & 61.16 & 54.72 & 61.47  \\ 
Biaff&   67.06 & 68.22 & 60.39 & 55.60 & 61.40 \\ 
\bottomrule
\end{tabular}
\caption{\small Our model with different prediction modules.}
\label{tab:prediction_ablation}
\end{table}

\subsection{Results on SemEval 2018 Task 7}

The results on SemEval 2018 Task 7 are shown in Table~\ref{tab:semeval2018}.
Our Entity-Aware BERT$_{\textrm{SP}}$ gives comparable results to the top-ranked system~\cite{rotsztejn-etal-2018-eth} in the shared task, with slightly lower Macro-F1, which is the official metric of the task, and slightly higher Micro-F1.
When predicting multiple relations in one-pass, we have 0.9\% drop on Macro-F1, but a further 0.8\% improvement on Micro-F1.
Note that the system~\cite{rotsztejn-etal-2018-eth} integrates many techniques like feature-engineering, model combination, pre-training embeddings on in-domain data, and artificial data generation, while our model is almost a direct adaption from the ACE architecture.

On the other hand, compared to the top single-model result~\cite{luan-etal-2018-uwnlp}, which makes use of additional word and entity embeddings pre-trained on in-domain data, our methods demonstrate clear advantage as a single model. 

\begin{table}[!t]
\small
\centering
\begin{tabular}{lcc}
\toprule
\multirow{2}{*}{\textbf{Method}} &
\multicolumn{2}{c}{\textbf{Averaged F1}} \\ 
& \textbf{Macro}& \textbf{Micro} \\
\midrule
\multicolumn{3}{c}{\underline{\emph{Top 3 in the Shared Task}}}\\
\cite{rotsztejn-etal-2018-eth} & \bf 81.7 & 82.8\\
\cite{luan-etal-2018-uwnlp} & 78.9 & -\\
\cite{nooralahzadeh-etal-2018-sirius} & 76.7 & -\\
\midrule
Ours (single-per-pass) & 81.4 & 83.1 \\
Ours (multiple-per-pass) & 80.5 & \bf 83.9 \\
\bottomrule
\end{tabular}
\caption{\small Results on SemEval 2018 Task 7, Sub-Task 1.1.}
\label{tab:semeval2018}
\end{table}

\subsection{Additional SRE Results}
We conduct additional experiments on the relation classification task, SemEval 2010 Task 8, to compare with models developed on this benchmark.
From the results in Table \ref{tab:semeval2010}, our proposed techniques also outperforms the state-of-the-art on this single-relation benchmark.

\begin{table}[!t]
\small
\centering
\begin{tabular}{lc}
\toprule
\multicolumn{1}{l}{\textbf{Method}} & \textbf{Macro-F1} \\ \midrule
Best published result \cite{wang2016relation} & 88.0\\
\midrule
BERT out-of-box                      & 80.9         \\
Entity-Aware BERT & 88.8         \\
BERT$_{\textrm{SP}}$ & 88.8         \\
Entity-Aware BERT$_{\textrm{SP}}$ & \bf 89.0         \\ 
\bottomrule
\end{tabular}
\caption{\small Additional Results on SemEval 2010 Task 8.}
\label{tab:semeval2010}
\end{table}

On this single relation task, the out-of-box BERT achieves a reasonable result after fine-tuning. Adding the entity-aware attention gives about 8\% improvement, due to the availability of the entity information during encoding.
Adding structured prediction layer to BERT (i.e., BERT$_{\textrm{SP}}$) also leads to a similar amount of improvement.
However, the gap between BERT$_{\textrm{SP}}$ method with and without entity-aware attention is small.
This is likely because of the bias of data distribution: the assumption that only two target entities exist, makes the two techniques have similar effects.

\section{Conclusion}
In summary, we propose a first-of-its-kind solution that can simultaneously extract multiple relations with one-pass encoding of an input paragraph for MRE tasks.
With the proposed structured prediction and entity-aware self-attention layers on top of BERT, we achieve a new state-of-the-art results with high efficiency on the ACE 2005 benchmark.
Our idea of encoding a passage regarding multiple entities has potentially broader applications beyond relation extraction, e.g., entity-centric passage encoding in question answering~\cite{song2018exploring}. In the future work, we will explore the usage of this method with other applications.

\bibliography{acl2019}
\bibliographystyle{acl_natbib}

\end{document}